\documentclass[letterpaper, 10 pt, conference]{ieeeconf}  %

\IEEEoverridecommandlockouts                              %

\usepackage{graphicx}
\graphicspath{{images/}}
\usepackage{tensor}
\usepackage{subfig}

\usepackage{mathtools}
\usepackage{algorithm}
\usepackage[noend]{algpseudocode}
\algrenewcommand\algorithmicrequire{\textbf{Input:}}
\algrenewcommand\algorithmicensure{\textbf{Output:}}
\newcommand*\Let[2]{\State #1 $\gets$ #2}
\algnewcommand\algorithmicforeach{\textbf{for each}}
\algdef{S}[FOR]{ForEach}[1]{\algorithmicforeach\ #1\ \algorithmicdo}
\DeclarePairedDelimiterX{\norm}[1]{\lVert}{\rVert}{#1}

\usepackage{nameref}
\usepackage{outlines}
\usepackage{booktabs}
\usepackage[pagebackref=true,breaklinks=true,letterpaper=true,colorlinks,bookmarks=false]{hyperref} %
\usepackage{float}
\usepackage{amssymb}
\usepackage{amsmath}
\usepackage{bbm}

\usepackage{color}
\newcount\Comments  %
\definecolor{darkgreen}{rgb}{0,0.5,0}
\definecolor{darkred}{rgb}{0.7,0,0}
\definecolor{teal}{rgb}{0.3,0.8,0.8}
\definecolor{orange}{rgb}{1.0,0.5,0.0}
\definecolor{purple}{rgb}{0.8,0.0,0.8}
\newcommand{\kibitz}[2]{\ifnum\Comments=1{\textcolor{#1}{\textsf{\footnotesize
				#2}}}\fi}

\title{\LARGE \bf
Learning Optimal Decision Making for an Industrial\\Truck Unloading Robot using Minimal Simulator Runs 
}

\author{Manash Pratim Das$^{1}$, Anirudh Vemula$^{1}$, Mayank Pathak$^{2}$, Sandip Aine$^{1}$, Maxim Likhachev$^{1}$%
\thanks{M. P. Das, A. Vemula, S. Aine and M. Likhachev are with the Robotics Institute, Carnegie Mellon University, Pittsburgh, PA, USA 15213. Email
        {\tt\small \{mpratimd,avemula1,asandip,mlikhach\}@cmu.edu}}%
\thanks{M. Pathak is with Honeywell, Pittsburgh, PA, USA 15205. Email
	{\tt\small mayank.pathak@honeywell.com }}%
\thanks{This work is funded by Honeywell and supported by National Robotics Engineering Center (NREC), Carnegie Mellon University, Pittsburgh, USA. 
}%
}

\begin{document}

\maketitle
\thispagestyle{empty}
\pagestyle{empty}

\begin{abstract}

Consider a truck filled with boxes of varying size and unknown mass and an industrial robot with end-effectors that can unload multiple boxes from any reachable location. In this work, we investigate how would the robot with the help of a simulator, learn to maximize the number of boxes unloaded by each action. Most high-fidelity robotic simulators like ours are time-consuming. Therefore, we investigate the above learning problem with a focus on minimizing the number of simulation runs required. The optimal decision-making problem under this setting can be formulated as a multi-class classification problem. However, to obtain the outcome of any action requires us to run the time-consuming simulator, thereby restricting the amount of training data that can be collected. Thus, we need a data-efficient approach to learn the classifier and generalize it with a minimal amount of data. A high-fidelity physics-based simulator is common in general for complex manipulation tasks involving multi-body interactions. To this end, we train an optimal decision tree as the classifier, and for each branch of the decision tree, we reason about the confidence in the decision using a Probably Approximately Correct (PAC) framework to determine whether more simulator data will help reach a certain confidence level. This provides us with a mechanism to evaluate when simulation can be avoided for certain decisions, and when simulation will improve the decision making. For the truck unloading problem, our experiments show that a significant reduction in simulator runs can be achieved using the proposed method as compared to naively running the simulator to collect data to train equally performing decision trees. 

\end{abstract}

\section{Introduction}
\begin{figure}[t!]
	\centering
	\includegraphics[width=0.5\columnwidth]{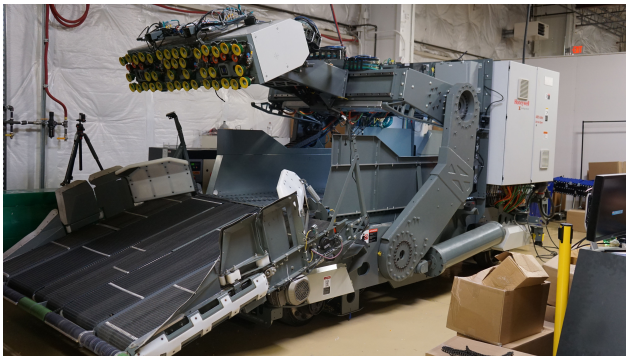}%
	\includegraphics[width=0.5\columnwidth]{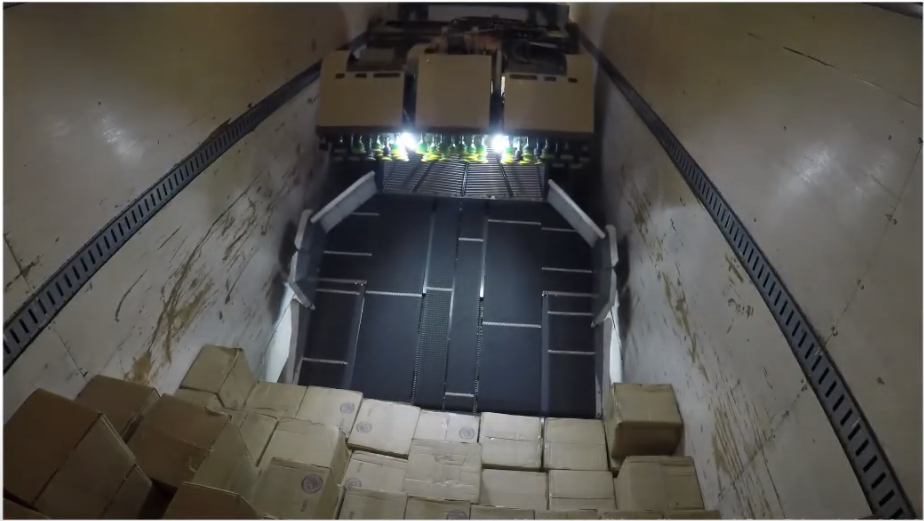}
	\caption{\small \textbf{Left}: The industrial truck unloading robot. \textbf{Right}: Boxes inside a truck with the robot facing them.}
	\vspace{-6mm}
	\label{fig:motivating-problem}
\end{figure}

Many robotics applications require planning and decision making based on what the robot observes in order to complete a task. In this article we study the problem of robotic truck unloading, where the task is to empty the truck by unloading all the boxes. \autoref{fig:motivating-problem} shows an industrial truck unloading robot and cardboard boxes inside the truck that needs to be unloaded. The robot has two end effectors, one is fixed to the base of the robot and can sweep boxes from the floor, and another suspended with an arm and can pick boxes using a plunger mechanism. The task of emptying the truck can be broken down into small sub-tasks like picking boxes from a certain location or sweeping boxes from the floor. This problem involves both task planning and motion planning{\tiny }. We exploit the fact that our problem allows us to perform task planning and motion planning independently, and hence in this paper, we focus only on the task planning. 

\section{Related Works}
\label{related-works}
The truck unloading problem has also been studied in our previous work \cite{islam2020planning,kim2019pomhdp}, and by Doliotis et.al.\ in \cite{doliotis20163d}. Our previous work looked at this problem as a sequential decision making problem. Whereas in this work, we formulate it as an immediate reward maximization problem. Precisely, given a certain configuration of boxes (scenarios), our problem is to determine the best task among a discrete set of tasks (such as picking from certain locations and sweeping at a certain depth) to unload the maximum number of boxes (along with other objectives such as minimizing the number of boxes damaged). The number of boxes can be in the order of hundreds, and their physical properties such as shape, weight and surface material are unknown and can vary widely. Thus, standard Task and Motion Planning (TAMP) approaches \cite{TAMP}, as well as Model Predictive Control \cite{mpc} which uses Lagrangian state representation (position of each box) to solve this problem would be impractical. The major reasons being that analytical models to forward simulate such complex interactions would be computationally very expensive, and obtaining the ``full-state'' of each box in the pile along with their physical properties is not realistically possible. Therefore, the task needs to be solved usually with compressed state representation of the whole pile of boxes as realistically perceivable by standard sensors. This introduces partial observability of the complete state and singularities where the inverse map from compressed state to a full state state is not unique. As shown in our previous work \cite{islam2020planning}, to handle uncertainty, a standard approach would be to compute a probabilistic distribution of the compressed state (belief-state), and formulate the planning problem as solving a Belief Markov Decision Process \cite{kaelbling1998planning}. However, we observed that the amount of data required for credit assignment and sequential reasoning is relatively very large. Also due to \textit{curse of dimensionality} and \textit{curse of history}, it is practically infeasible to determine an optimal sequence of actions which will empty the truck. 

In this paper, we therefore focus on the immediate reward maximization objective and ensure that it is practically feasible to find the optimal decision at each step. However, this assumes that the goal (empty trailer) is reachable from every state. We also assume that it is possible to reset the simulator to a box configuration such that we can record how each action performs on the configuration. These are the only two assumptions made by the proposed approach. With these assumptions, the na\"ive method to solve this problem is to simulate all tasks for a wide variety of scenarios in order to observe the best action for each scenario, and to train a classifier based on this data. However, the simulation time and computation required to collect such an extensive dataset would make the na\"ive method unfeasible, as each high-fidelity simulation run needs to model all the physical interactions between the boxes (which can go up to 1000 in number) and the robot. Thus, we are limited by the maximum number of simulations we can perform feasibly. Under these limitations, a data-efficient approach to train the classifier is required.

In \autoref{sec:main-algo} we present our main contribution which is an online-algorithm that iteratively trains a classifier while building a dataset by selectively running the simulator. The key idea behind this work is that we can reject running the simulator for (easy) scenarios where the robot is already confident about it's decision. This can help focus the resources on more difficult scenarios. In contrast to Active Learning \cite{settles2012active} methods, new scenarios are only being given to us by a transition function (see \autoref{alg:ITRS}), and it is not possible to actively setup a physically valid scenario. Our problem setup might look similar to the Contextual version of the Multi-Armed Bandit problem \cite{katehakis1987multi}, where you only get to observe the result of one single action for a scene. However, they are not similar, because in our setup, it is possible to reset the simulator and observe the result of multiple actions for any scene. Setups such as ours are common in the robotics domain involving a simulator.

\section{Problem Formulation}
\label{sec:problem-formulation}
Let an arrangement of boxes inside a trailer be represented as a state $s\in S$, where let $S$ denote a set of all such box arrangements (box states) and $d_S$ denote the distribution over these states. The perception system on the robot would generate a 3D voxel grid $v_s \in \mathcal{V}$ for that state $s$. Now, suppose there exists a method $\Phi: \mathcal{V} \rightarrow X$ to generate fixed size features from such voxel grids (more details in \autoref{sec:bin-features}). The robot can only use these features, derived from perception data of the world to make its decisions. Thus, for decision making, we need a policy $\pi \in \Pi, \pi : \mathcal{V} \rightarrow A$, which gives us the high-level action the robot should execute for the a feature representation of the state. The action space $A$ can be continuous, but in this work, we are interested in a finite action space due to two reasons: 1) a finite action space makes it easier to derive theoretical guntees of the form we discuss in \autoref{sec:bandits}, 2) in practice one may find a reasonable discretization of a bounded continuous space. Further, $\pi$ can be considered as a classifier where $A$ is the set of all classes. In the proposed method we use Optimal Sparse Decision Trees (OSDT) \cite{hu2019optimal}. Thus, $\Pi$ is the class of decision trees considered in OSDT. Finally, the goal is to find a policy $\pi^*\in \Pi$ which maximizes immediate reward using minimal execution of the resource consuming simulator. For context, our simulation environment on CoppeliaSim \cite{rohmer2013coppeliasim}, uses all 6 vCPUs, 2.5 GHz processors on an AWS cloud machine and takes around 10 minutes to simulate one action in any state. Note that, we train the classifier on data that we obtain from the simulator and we evaluate it in simulation itself. Bridging the sim-to-real gap is out of the scope of this paper.

\subsection{Preliminaries}
\label{sec:prelims}
A decision tree is characterized by its partition of the feature space and the decision taken by the leaves of the tree for each of these partitions. Let $\Pi_\text{s}$ denote the class of decision trees with the same structure / partition of feature space but with different decisions at its leaves.

Consider a leaf node $l$ of a decision tree. As, described earlier, the decision tree partitions the feature space and a leaf corresponds to one such partition. Further, let $S_l$ and $d_{S_l}$ denote the set and distribution respectively of world states whose feature representation lies in the partition corresponding to the node $l$. Let $R(s,a_i) \in [0,1]$ denote a random variable for the reward received on applying action $a_i \in A$ on state $s$. In the truck-unloading problem, consider this reward to be a combination of interest variables such as the unload rate achieved, boxes dropped or boxes damaged upon executing an action $a_i$ on the state $s$. For simplicity of notation, let $R_l(a_i) = \mathbb{E}_{s \sim d_{S_l}}[ R(s,a_i) ]$ denote the \textbf{true expected reward} for action $a_i$ on the states that fall in node $l$. Note that $R_l(a_i)$ is unknown, and is still a random variable since we assume randomness in the rewards obtained from executing an action. Now, for each leaf node, where a decision is taken, if we somehow had a way to know these true expected rewards, we would have picked the action that has maximum true expected reward, and hence would have found the optimal policy in $\Pi_\text{s}$. However, we can only get empirical estimate for these true expected rewards, and it might require infinitely huge dataset to determine the truly best optimal action. Instead, in practice it would be sufficient to find an  $\epsilon$-optimal action for each leaf of the decision tree. An $\epsilon$-optimal action $a'$, for any node $l$, is an action for which the condition $\{\mathbb{E}[R_l(a')] > \mathbb{E}[R_l(a^*)] - \epsilon\}$ holds true with a high probability, where $a^* = \arg\max_{a_i\in A} \mathbb{E}[R_l(a_i)]$. Note that we discuss the bounds only in a Probably Approximately Correct (PAC) setting as described in the following sections.

Given a decision tree structure as in $\Pi_\text{s}$, in~\autoref{sec:bandits}, we will discuss how to determine if we have found an $\epsilon$-optimal action for each of the leaf nodes $l$. We do this by eliminating all actions that under the PAC setting, cannot be the $\epsilon$-optimal action. We would then have a systematic way of running the simulator to obtain more data to resolve among the non-eliminated actions. 

\section{Re-visiting Multi-Armed Bandits}
\label{sec:bandits}
We will now discuss algorithms that will help us determine the $\epsilon$-optimal action in a PAC setting for any leaf node $l$. The class of decision trees $\Pi$ we consider are deterministic, in that, given a world state $s$, its feature representation would deterministically fall into a partition / leaf node. Thus, given a decision tree structure $\Pi_\text{s}$, the set $S_l$ is fixed. So, the decision at each leaf $l$ is affected by only the states $S_l$, and the rewards $R_l(a_i)$ received at that leaf for each action $a_i$. In other words, the random variable $\mathbb{E}[R_l(a_i)]$ that is independent across all the leaf nodes $l$. Thus, we consider the decision making problem at each leaf as a independent multi-armed bandit problem (MABPs), and we derive an algorithm that can be used while training the decision tree to eliminate actions (prevent executing of simulation for those actions).

Without loss of generality, let us consider a leaf node $l$, and drop the subscript `$l$' for
notational simplicity. Let $\hat{r}_i = \frac{1}{\tau}\sum_{j=1}^{\tau}R(s_j,a_i)$ be the \textbf{average empirical reward} for action $a_i$ for the states $s_j, j=1,\ldots,\tau$ in training data that fall into the feature partition corresponding to the leaf node. Similarly, let $r_i = R_l(a_i)$. Here executing an ``action'' in the simulator corresponds to sampling an ``arm'' in the MABP literature. Thus, the problem is to choose the best action $a'\in A$ out of $n=|A|$
total action. First, we discuss a Naive Algorithm (\texttt{NV}) (\autoref{alg:naive}) as presented in \cite{even2002pac}, which gives the sample complexity required to determine with a probability of ($1-\delta$) an $\epsilon$-optimal arm. Such algorithms are termed as ($\epsilon,\delta$)-\texttt{PAC} algorithms.

\begin{algorithm}[t!]
	\caption{Naive ($\epsilon,\delta$)-\texttt{PAC} algorithm
		\label{alg:naive}}
	\begin{algorithmic}[1]
		\Require{Number of arms $n$, suboptimality $\epsilon$, probability $\delta$}

		\State For every arm $a_i\in A$: Sample it $\tau = \frac{4}{\epsilon^2}\log\left(\frac{2n}{\delta}\right)$ times
		\State Let $\hat{r}_i$ be the empirical average reward of arm $a_i$ from the samples collected
		\State \Return{The ($\epsilon,\delta$)-best arm $a' = \arg\max_{a_i \in A}\{\hat{r}_i\}$}
	\end{algorithmic}
\end{algorithm}
According to this algorithm, we can find an ($\epsilon,\delta$)-action for a leaf, only after running the simulator for $\frac{4n}{\epsilon^2}\log\left(\frac{2n}{\delta}\right)$ times on states that are partitioned into this leaf.

\autoref{alg:naive} is very wasteful as it waits until the required sample complexity is reached. For example, if $n=7, \epsilon=0.1, \delta=0.05$, the number of simulations required per leaf is 15778 ($\sim547.8$ days of simulation for a tree with 5 leaves). A key insight for our domain is that, evaluating whether we have found an ($\epsilon,\delta$)-action is computationally negligible compared to executing expensive simulation. Successive Elimination Algorithm (\texttt{SE}) (\autoref{alg:successive-elimination}) \cite{even2002pac}, an iterative algorithm, exploits this idea and executes each non-eliminated action one additional time per iteration before evaluating which actions can be eliminated. It is a ($0,\delta$)-\texttt{PAC} algorithm as it runs until the ($0,\delta$)-action is found. A ($0,\delta$)-\texttt{PAC} algorithm can be modified to a ($\epsilon,\delta$)-\texttt{PAC} algorithm by stopping early when $\tau_t = \frac{4}{\epsilon^2}\log\left(\frac{2n}{\delta}\right)$ (based on \autoref{alg:naive}) and returning the arm $a' = \arg\max_{a_i \in \chi_t}\{\hat{r}_{i,t}\}$.

\begin{algorithm}[t!]
	\caption{($0,\delta$)-\texttt{PAC} Successive Elimination
		\label{alg:successive-elimination}}
	\begin{algorithmic}[1]
		\Require{$n$, $\delta$, existing count of samples $\tau$ for each arm}
		\Let{$\chi_\tau$}{$\{a_1,a_2,\ldots,a_n\}$}
		\State Compute $\hat{r}_{i, \tau}$ for all arms based on $\tau$ samples
		\While{$|\chi_\tau| > 1$}
			\Let{$\epsilon_\tau$}{$\sqrt{ \frac{2}{\tau}\log(\frac{4\tau^2n}{\delta}) } $}
			\Let{$\chi_{\tau+1}$}{$\chi_\tau \backslash \{a_i \in \chi_\tau \: | \: \max_{j\in \chi_\tau} \hat{r}_{j,\tau} - \hat{r}_{i,\tau} > 2\epsilon_\tau\}$}
			\State Sample each arm in $\chi_{\tau+1}$, and compute $\hat{r}_{i,\tau+1}$
			\Let{$\tau$}{$\tau+1$}
		\EndWhile
		\State \Return{The only arm left in $\chi_\tau$}
	\end{algorithmic}
\end{algorithm}
In \autoref{alg:successive-elimination}, note the average empirical reward $\hat{r}_{i,\tau}$ is evaluated at each iteration $\tau$, and the algorithm can be started even with existing data. The most important thing to note here is that at every iteration $\tau$, the actions that are left in the set $\chi_\tau$ are all sampled equal number of times $\tau+1$, and elimination is based on this fact. Actions which were eliminated were executed $\leq\tau$ times. If we have an action which is clearly better than all of the rest, ($0,\delta$)-\texttt{PAC} \texttt{SE} has the potential to eliminate all that actions except the best one using lesser samples than ($\epsilon,\delta$)-\texttt{PAC} \texttt{NV}. ($\epsilon,\delta$)-\texttt{PAC} \texttt{SE} might require even lesser samples as it needs to eliminate only those actions which cannot form an $\epsilon$-optimal action (with high probability). Thus, depending upon the structure of the problem, it may be possible to exploit the fact that we can clearly identify some actions which are worse.

\section{Successive Elimination with Decision Trees}
In \autoref{sec:bandits}, we discussed the ($0,\delta$)-\texttt{PAC} \texttt{SE} algorithm in a setting where $\Pi_\text{s}$ was fixed. However, as we will discuss in \autoref{sec:main-algo}, we will iteratively update $\Pi_\text{s}$ by training a new decision tree when new simulation data is available. Suppose some actions were eliminated, and the rest were sampled for a given $\Pi_\text{s}$. Now, when $\Pi_\text{s}$ is updated, the feature partitions will change and would no longer capture the same states. Note that, now for some state-action pair, we might not have rewards, as that action might have been eliminated in some previous version of $\Pi_\text{s}$. As a result, each leaf might have actions which are sampled different number of times, and elimination as performed in ($0,\delta$)-\texttt{PAC} \texttt{SE} (\autoref{alg:successive-elimination}) can no longer be applied here. To this end, we present a modified version of \texttt{SE}, which can handle non-uniform sampling (\autoref{alg:nu-successive-elimination}). Further, eliminations are no longer persistent. In other words, with variable $\Pi_\text{s}$, we can no longer assume that an action eliminated for one leaf cannot become the ($\epsilon,\delta$)-action in some other leaf, before and after updating $\Pi_\text{s}$. Therefore, let us no longer think about eliminating actions until the best action is found, rather think about executing simulation for only those actions which may be the best action (with high probability), and reject executing simulation for the rest of the actions. We therefore, define a sub-routine \textsc{SelectActions} used by the Non-Uniform Successive Elimination (\texttt{NUSE}) (\autoref{alg:nu-successive-elimination}). Let us now consider a iteration $t$, where the decision tree structure is defined by $\Pi_{\text{s},t}$. Consider any leaf $l$, and let $\tau_{i,t}$ denote the number of data samples for action $i$, captured by $l$ in that iteration $t$. Therefore, let $\hat{r}_{i,t}$ denote the average empirical reward based on $\tau_{i,t}$ samples. \autoref{alg:nu-successive-elimination} is a ($0,\delta$)-\texttt{PAC} algorithm and we present the proof in \nameref{sec:appendix}. Additionally, stopping early when $\tau_{i,t} \geq \frac{4}{\epsilon^2}\log\left(\frac{2n}{\delta}\right)$ for all $a_i \in \chi_t$, gives us ($\epsilon,\delta$)-\texttt{PAC} \texttt{NUSE} version of the algorithm (proof in \nameref{sec:appendix}).

\begin{algorithm}[t!]
	\caption{Non-Uniform Successive Elimination
		\label{alg:nu-successive-elimination}}
	\begin{algorithmic}[1]
		\Require{$n$, $\delta$, \{$\tau_{i,t}\}, t$}
		\Function{Main}{$n$, $\delta$, \{$\tau_{i,t}\}, t$}
			\Let{$\chi$}{$\{a_1,a_2,\ldots,a_n\}$}
			\State Compute $\hat{r}_{i, t}$ based on $\tau_{i,t}$ samples for all arms $i$
			\While{true}
				\Let{$\chi_{t}$}{\textsc{SelectActions}($\{\tau_{i,t}\}, \{\hat{r}_{i,t}\}$)}
				\If{$|\chi_t| = 1$}
					\State \Return{The only arm left in $\chi_t$}
				\Else
					\State Sample any arm in $\chi_{t}$, any number of times
					\State Update $\tau_{i,t+1}$ and compute $\hat{r}_{i,t+1}$ for all $i$
					\Let{$t$}{$t+1$}
				\EndIf
			\EndWhile
		\EndFunction
		\Function{SelectActions}{$\{\tau_{i,t}\}, \{\hat{r}_{i,t}\}$}
			\Let{$\epsilon_{i,t}$}{$\sqrt{ \frac{2}{\tau_{i,t}}\log(\frac{4t^2n}{\delta}) } $ for all $i$}
			\Let{$\chi^c$}{$\{a_i \in \chi \: | \: \max_{j\in \chi} \hat{r}_{j,t} - \hat{r}_{i,t} > 2\max_i\epsilon_{i,t}\}$}
			\Let{$\chi_{t}$}{$\chi \backslash \chi^c$}
			\State \Return $\chi_{t}$
		\EndFunction
	\end{algorithmic}
\end{algorithm}

\begin{algorithm}[t!]
	\caption{Iterative Training With Rejection Sampling
		\label{alg:ITRS}}
	\begin{algorithmic}[1]
		\Require{$n$, $\delta$, sub-optimality $\epsilon$, initial dataset $\mathcal{D}_t$}
		\Function{Main}{}
		\Let{$\chi$}{$\{a_1,a_2,\ldots,a_n\}$} \Comment{Set of all actions}
		\Let{$t$}{$|\mathcal{D}_t|/n$}
		\Let{$s_t$}{Initial state in the Simulator}
		\Let{$L_t$}{1} \Comment{An initial value $>0$}
		\While{$L_t > 0$ and $t < T$} \Comment{$T$ is the max iteration}
		\Let{$E_{t}$}{Train Encoder network using $\mathcal{D}_{t}$}
		\Let{$\hat{X}_{t}$}{$E_{t}(X_{t}\in \mathcal{D}_{t})$}
		\Let{$\pi_{t}$}{Train OSDT using $\mathcal{D}_{t}$ and $\hat{X}_{t}$}
		\Let{$v_t$}{3D voxel of $s_t$ from perception}
		\State $x_t\gets \Phi(v_t) \qquad \hat{x}_t \gets  E_t(x_t)$
		\Let{$l$}{leaf node in $\pi_t$ that captures $\hat{x}_t$}
		\Let{$\{\tau_{i,t}\}, \{\hat{r}_{i,t}\}$}{\textsc{LeafInformation}($\pi_t,l,\mathcal{D}_t$)}
		\Let{$\chi_t$}{\textsc{SelectActions}($\{\tau_{i,t}\}, \{\hat{r}_{i,t}\}$)}
		\Let{$\mathcal{D}_{t+1}$}{$\mathcal{D}_t$}
		\Let{$\hat{S}, \chi_t$}{$\emptyset, \emptyset$}
		\If{$\min\{\tau_{i,t}\} < \frac{4}{\epsilon^2}\log\left(\frac{2n}{\delta}\right)$ and $|\chi_t|>1$}
		\ForAll{$a_i \in \chi_t$}
		\Let{$R(s_t,a_i),s_{i,t+1}$}{\textsc{Simulate}($s_t,a_i$)}
		\Let{$d$}{$(x_t, a_i, R(s_t, a_i))$}
		\Let{$\mathcal{D}_{t+1}$}{$\mathcal{D}_{t+1} \cup d$}\Comment{Extend Dataset}
		\Let{$\hat{S}$}{$\hat{S}\cup\{s_{i,t+1}\}$}
		\EndFor				
		\EndIf
		\Let{$s_{t+1}, a_r, r_r$}{\textsc{TransitionState}($\hat{S}, \chi_t, s_t$)}
		\Let{$d$}{$(x_t, a_r, r_r)$}
		\Let{$\mathcal{D}_{t+1}$}{$\mathcal{D}_{t+1} \cup d$}\Comment{Extend Dataset}
		\Let{$L_{t+1}$}{\textsc{RemainingLeaves}($\pi_t$)}
		\Let{$t$}{$t+1$}
		\EndWhile
		\State \Return{$\pi_t$ as $\pi^*$}
		\EndFunction
		\Function{RemainingLeaves}{$\pi$}
		\State \Return{Number of leaves in $\pi$ where an ($\epsilon,\delta$)-action is yet to be found}
		\EndFunction
		\Function{LeafInformation}{$\pi_t,l,\mathcal{D}_t$}
		\State \Return{sample count and avg. expected rewards for each action based on the data captured by leaf $l$ in $\pi_t$}
		\EndFunction
		\Function{TransitionState}{$\hat{S}, \chi_t, s_t$}
		\State{Any function that can choose an action $a_r$ to be applied on $s_t$, and to obtain new state $s_{r,t+1}$ with the associated reward $R(s_t,a_r)$ }
		\State \Return{$s_{r,t+1}, a_r, R(s_t,a_r)$}
		\EndFunction
	\end{algorithmic}
\end{algorithm}
\setlength{\textfloatsep}{0.1cm}

\section{Iterative Training with Rejection Sampling}
\label{sec:main-algo}
\autoref{alg:ITRS} presents the proposed algorithm. It trains an optimal decision tree $\pi^*$, starting from an initial dataset $\mathcal{D}_t$ (at least $|A|$ reward samples for each action in a state), and extending this dataset by collecting more simulation data only when required according to ($\epsilon,\delta$)-\texttt{PAC} \texttt{NUSE}. However, we first discuss the features and how we train with a sparse dataset.

\subsection{Task-Relevant Binary Features}
\label{sec:bin-features}
In \autoref{sec:problem-formulation}, we described $\Phi : \mathcal{V} \rightarrow X$, a function that generates fixed size features from 3D voxel grid $v \in \mathcal{V}$. The feature space $X$ can be continuous, and can be of any dimension. However, the OSDT \cite{hu2019optimal} algorithm works only with binary features. Thus to train OSDT for an arbitrary feature space $X$, first we have to come up with a map $\Psi:X\rightarrow \hat{X}$ (feature engineering), where $\hat{X}=\{0,1\}^m$ is a $m$-dimensional binary space, and then train OSDT using the binary features. The inconvenience of finding $\Psi$ is a price we have to pay to obtain some guarantees on optimality. Searching for a globally optimal classifier in continuous feature space $X$ is NP-Hard. We argue that breaking the problem down into two parts 1) finding the optimal map $\Psi$, 2) searching for optimal decision tree in binary space $\hat{X}$ makes the problem more manageable, as regardless of the former also being NP-Hard, one can often hand-engineer very good map $\Psi$ for the task. The later problem too is NP-Hard, but \cite{hu2019optimal} provides an efficient method to search for the optimal tree. A na\"ive map from $X=[0,1]^k$ continuous space would be to first discretize each of the dimensions to say $j$ categories. Now this categorical space can be represented using a minimum of $m = \log_2(j^k)$ binary variables. The proposed method, also supports the use of Neural Networks Encoders (EncoderNet) to learn a dynamic map simultaneously with OSDT after every dataset extension. The EncoderNet can be used for compression to keep only the information required to make correct classification (task-relevant), and hence use far less than binary variables than that of naive ($\log_2(j^k)$). \autoref{sec:experiments} contains details on the EncoderNet and $\Phi$ we use for the truck unloading problem. 

\subsection{Training with Sparse Dataset}
\label{sec:training_with_sparse}
After executing an action $a_i$ on state $s$ we can compute the reward $R(s,a_i)$ for that action. Let $x = \Phi(v)$ be the feature representation of the perception data $v$ corresponding to the state $s$. We can form a tuple $(x,a_i,R(s,a_i))$. Our dataset $\mathcal{D}$ corresponds to the set of these tuples from multiple simulation runs on various states and with various actions. Note that for training a classifier with supervised-learning, for each data point, one needs a class label corresponding to the best action for the given state. Since our problem contains rewards, all misclassification are not equal and a reward(cost)-sensitive classification is a better approach. However with rejection sampling, the dataset may not contain rewards for all actions of all states, and thus generate a sparse dataset. We want the OSDT algorithm to 1) perform reward-sensitive classification, and 2) support sparse dataset. The OSDT algorithm is based on the Branch and Bound algorithm \cite{BnB}. It iterates through potential partitions of the binary feature space in search of the optimal partition. While evaluating a partition, for each leaf, it computes the empirically best action and the loss, which now have to be calculated differently to support the above two goals. Skipping details for brevity, instead of using misclassification error as a metric, we instead use the average empirical rewards. The action elimination in ($\epsilon,\delta$)-\texttt{PAC} \texttt{NUSE}) under the PAC-setting ensures that the dataset contains the reward for the best action of each state.

\subsection{Comments on the Algorithm}
First, we emphasize on Rejection Sampling. Note that in \autoref{alg:ITRS} (18-23), we reject simulation on a state $s_t$ if that leaf that captures the state has either found an $(\epsilon,\delta)$-action or eliminated all bad actions. Even when simulation is executed, it is only done for the actions which can be the optimal (selected by ($\epsilon,\delta$)-\texttt{PAC} \texttt{NUSE}). Next, note that the algorithm requires at-least $|A|$ data points in the initial dataset. It can easily be obtained by executing all actions at any state. Finally, note that new states are generated based on successor states given by the simulator (\textsc{TransitionState}). This is in line with the fact that in most robotic domains, the states are result of a process and often cannot be generated otherwise. This is one reason why we perform rejection sampling instead of Active Learning \cite{settles2012active}.

\section{Experiments and Results}
\label{sec:experiments}
\begin{figure}[t!]
	\centering
	\includegraphics[width=\columnwidth]{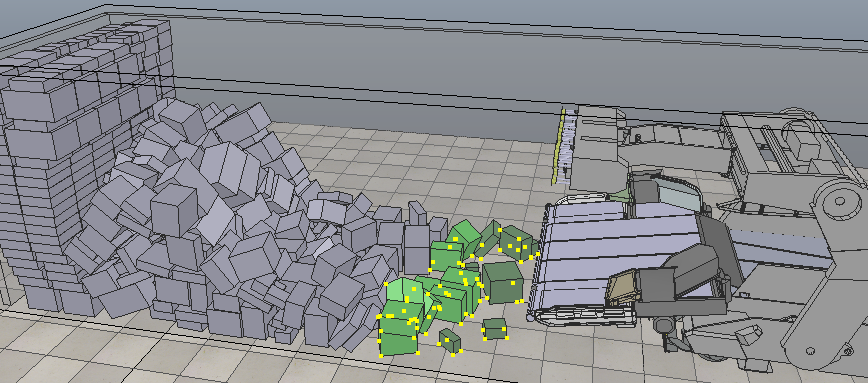}
	\caption{\small CoppeliaSim VREP simulator. The walls of the truck are made transparent for better visibility of the boxes inside.}
	\label{fig:simulator}
\end{figure}

Our experimental setup is as follows. Our simulator is based on the CoppeliaSim robot simulation platform. \autoref{fig:simulator} shows a simulation scene. The dimensions, masses and arrangement of the boxes we use during simulation closely resemble that in real-world operation. We use $\epsilon=0.45$ and $\delta=0.45$ in all our experiments.

\noindent\textbf{Pick vs Sweep problem:} As discussed in the sections above, in this paper, we look at the high-level decision making of the robot. Specifically, we look at the scenario where the robot has to decide 1) whether it should perform a ``Pick'' action, where it used the plungers at the end of its arm to grab boxes and pick them off from the truck, or 2) whether it should use the rollers at the base of the robot to ``Sweep'' up boxes from the floor of the robot. Once the high-level decision is made, we use other heuristic to determine the exact pick location or the sweep distance. We refer to our previous work \cite{islam2020planning} for details on how a high-level decision is executed for this Truck Unloading problem.

\begin{figure}[t!]
	\centering
	\includegraphics[width=\columnwidth]{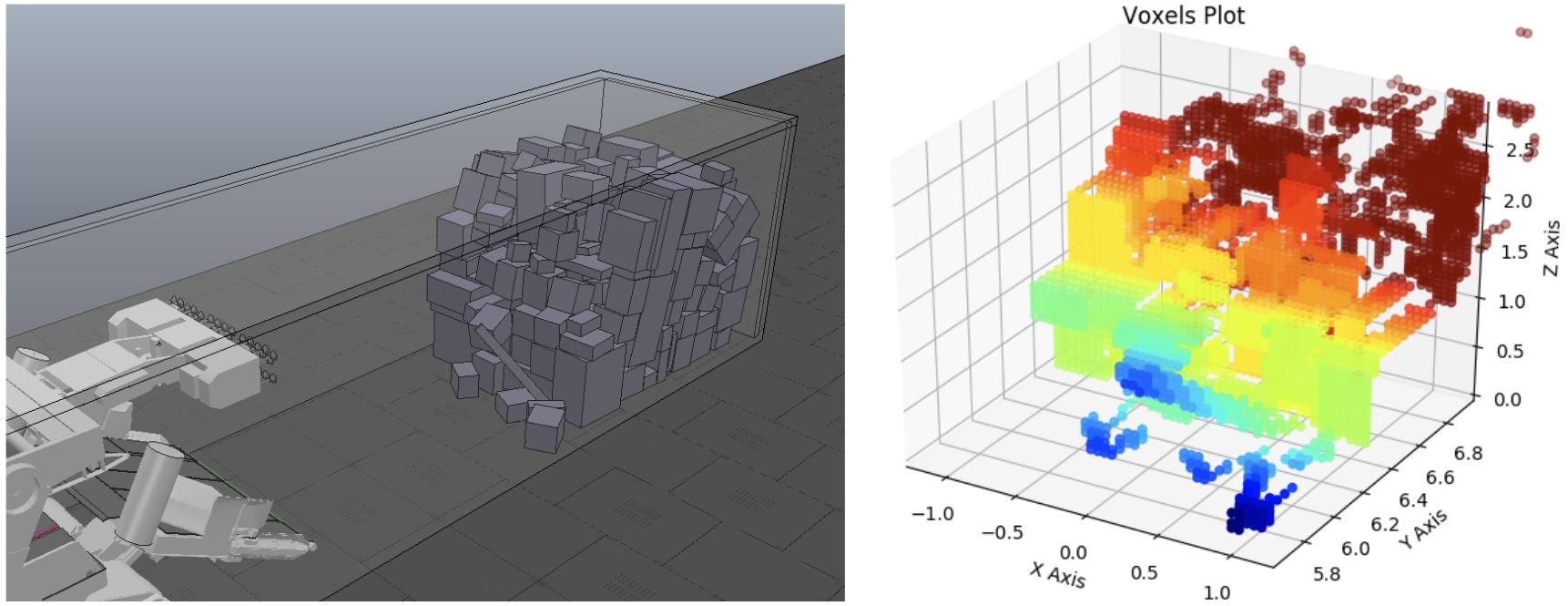}
	\caption{\small 3D occupancy grid $v$ (right) as provided by the perception system from the simulator state $s$ (left). The color of the voxel represent the height of the voxel.}
	\label{fig:perception}
\end{figure}

\begin{figure}[t!]
	\centering
	\includegraphics[width=\columnwidth]{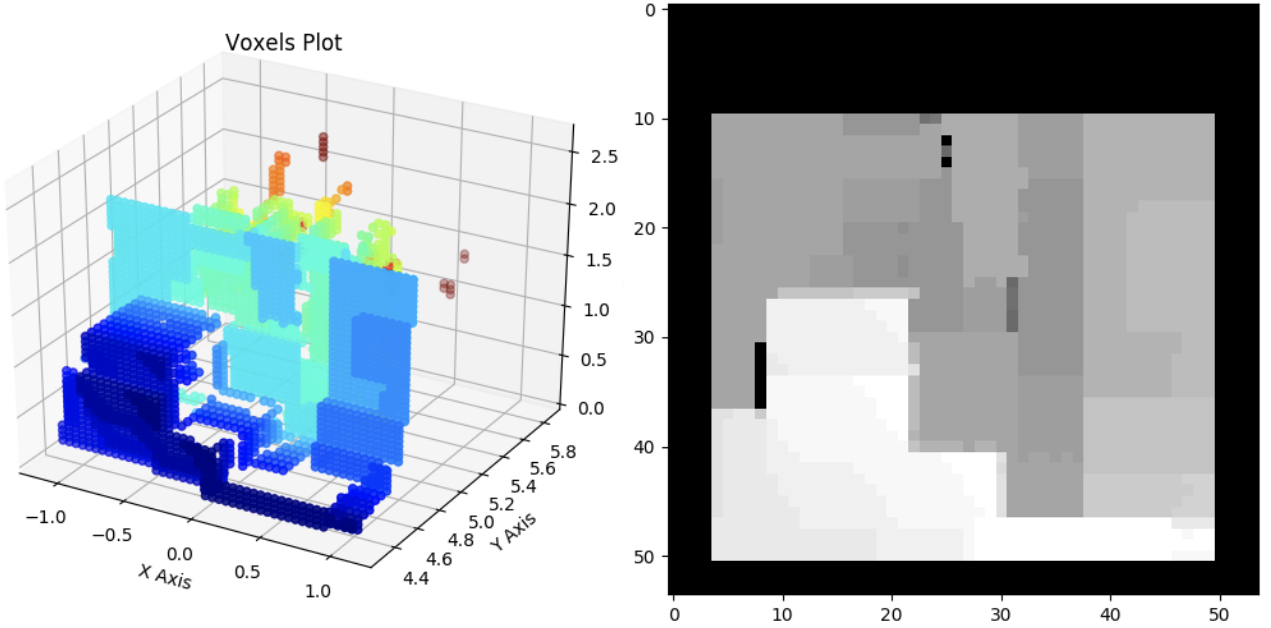}
	\caption{\small Depth Map generated by projecting the 3D occupancy grid}
	\label{fig:voxel-depth}
\end{figure}

\begin{figure}[h!]
	\centering
	\includegraphics[width=\columnwidth]{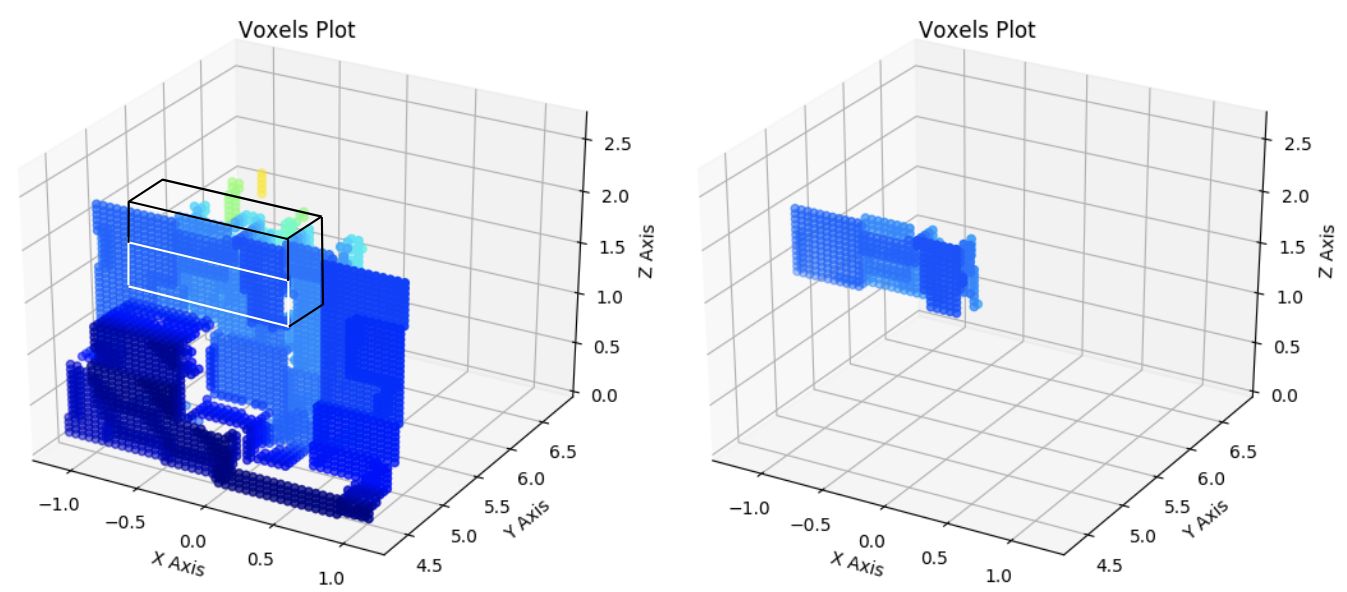}
	\caption{\small Visualizing the voxels that might be affected based on a Pick location.}
	\label{fig:ehcpbu}
\end{figure}

\begin{figure*}[t!]
	\centering
	\includegraphics[width=\textwidth]{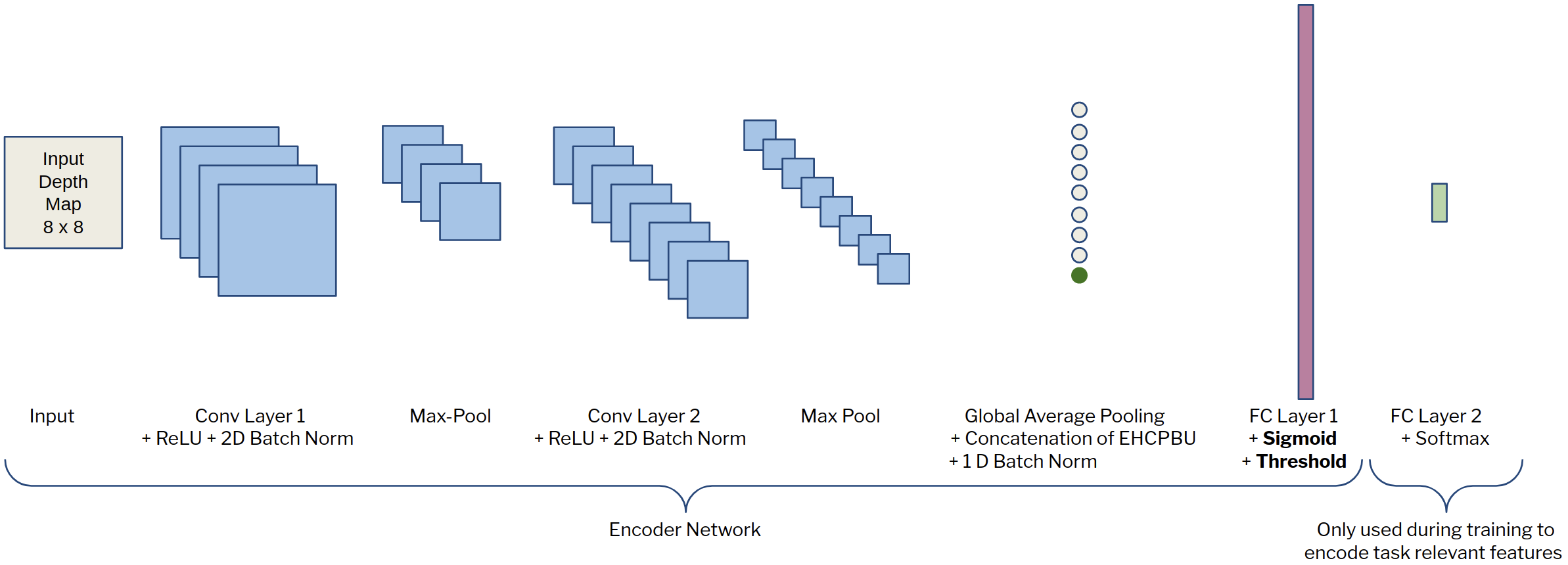}
	\caption{\small The Encoder Network Architecture.}
	\label{fig:encoder-net}
\end{figure*}

\begin{table*}[t!]
	\centering
	\caption{\small Classifier accuracy on standard test dataset with limited budget in training dataset}
	\label{table:accuracy}
	\begin{tabular}{|l|l|l|l|l|l|l|l|l|l|}
		\hline
		\textbf{Method} & \multicolumn{9}{c|}{\textbf{Simulation Call Budget}}                                                                                          \\ \hline
		& 200           & 300           & 400           & 500           & 600           & 700           & 800           & 900           & 1044          \\ \hline
		Naive           & 69.7          & 70.6          & 70.3          & 71.8          & \textbf{72.4} & 72.3          & 70.8          & 70.8          & 72.0          \\ \hline
		Random Rejector & 67.6          & 54.8          & 55.5          & 55.8          & 45.9          & 51.6          & 33.8          & 31.7          & 30.7          \\ \hline
		ITRS            & \textbf{70.8} & \textbf{71.2} & \textbf{70.6} & \textbf{72.1} & 71.8          & \textbf{72.9} & \textbf{71.5} & \textbf{73.0} & \textbf{74.1} \\ \hline
	\end{tabular}
\end{table*}

\noindent\textbf{Input Features}: We simulate perception sensors such that the high-level decision is based only based on simulated perception data and not ground-truth information from the simulator. This ensure that the model we use in simulation can also be used on the real robot. The perception system provides a 3D occupancy grid for the voxels \autoref{fig:perception}. To convert this voxel representation to a fixed size feature, we project the voxels on the plane perpendicular to the robot ($(x,z)$-plane as shown in \autoref{fig:perception}) and generate a gray-scale depth map of fixed size (54x54 pixels) \autoref{fig:voxel-depth}. We finally, down-sample the depth map to a size (8x8 pixels) to generate a small sized feature matrix. We arrive upon the down-sampled size empirically to determine the minimum size which can provide enough information required for decision making. Next we also extract what we call as EHCPBU features. Recall, that we use a hard-coded heuristic to determine the exact pick location once the robot decides to execute the ``Pick'' action. This information is available even before the decision is made, thus, we utilize it to enable more informed decision making. To that end, we compute the 3D volume of occupied voxels that might be affected based on the pick location and the plunger dimensions (\autoref{fig:ehcpbu}). Based on this 3D volume and a normalized box volume, we compute the estimated boxes that the Pick action might unload. Our observe a correlation of 0.525 between the EHCPBU estimate and the true number of boxes picked.

Hence, the method $\Phi$ in our case converts 3D voxels into depth-map and EHCPBU features. As described in \autoref{sec:bin-features}, we need binary features for the OSDT algorithm. \autoref{fig:encoder-net} shows the structure of the encoder network, which we use to convert the depth map and EHCPBU features to task-relevant binary features. Once trained, the output of the \texttt{FC Layer 1} after binary thresholding serves as the binary features for OSDT. 

\textbf{Baselines}:  As discussed in \autoref{related-works}, we do not compare our method against methods that fall into the category of Active Learning \cite{settles2012active} and Contextual Multi-Armed Bandits \cite{katehakis1987multi} which may look similar but do not share the same setup as ours. Moreover, while we use the immediate reward optimization problem formulation, it is not within the scope of this paper to compare against a sequential decision making problem formulation. Therefore, we only look at immediate-reward optimization baselines. In this regard, one of the most standard approach is to first generate data from the simulator and then train a classifier in an off-line manner. We call this the ``Na\"ive'' method. As compared to the ``Na\"ive'' method, the proposed ITRS \autoref{alg:ITRS} is an online-algorithm where we train the classifier while generating the dataset, by following an informed rejection scheme. We compare our method with another baseline which we call as ``Random Rejector'' which rejects running random $50\%$ of the actions for every state. We hope to capture the difference between a random rejection and an informed rejection using this baseline.

We run the following experiments:
\begin{enumerate}
	\item Performance of the classifier with limited budget on simulator calls
	\item Reduction in simulation time achieved
	\item Ablation study for the cost-sensitive classification
	\item In-depth analysis of ($\epsilon,\delta$)-\texttt{PAC} \texttt{NUSE}
\end{enumerate} 

\subsection{Performance of the classifier with limited budget on simulator calls}
\label{sec:limited-sims}
In this experiment, we analyze the case, when we fix the maximum number of simulation we can run. We evaluate the performance of the decision tree trained after the limit is reached. The ITRS and Random Rejector methods as they skip simulation runs, are allow to transition to new states independently. However, all three of the methods observe the same states and the same rewards for the actions in those states with the maximum overlap possible. OSDT algorithm with cost-sensitive loss function as described in \autoref{sec:training_with_sparse} is trained for all the three methods. \autoref{table:accuracy} shows the performance in terms of the accuracy with varying simulator limit. We observe that dataset collected with the ITRS method results in a more accurate classifier as compared to the ``Naive'' method in almost all of the cases. Interestingly, in the case of the Random Rejector method, we observe that the performance of the classifier decreased drastically as the dataset budget is increased. This may be contributed to the fact that, a random rejection would prevent the rewards for the true best action to be observed, in which case, the classifier would only have access to rewards for the worse actions during training. This effect gets compounded as the mistakes keep growing.

\subsection{Reduction in simulation time achieved}
\label{sec:reduction-in-sim}
\begin{figure}[h!]
	\centering
	\includegraphics[width=\columnwidth]{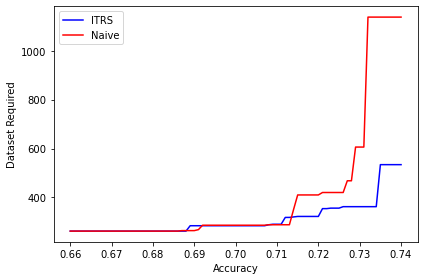}
	\caption{\small Size of the dataset required to achieve a certain classifier accuracy in the test dataset.}
	\label{fig:performance}
\end{figure}

 In this experiment, we count the number of simulator calls required by ITRS and the Na\"ive method to reach a certain testing accuracy of making the right decision in the Pick vs Sweep problem. In this experiment, we start initially with $4$ states, and we run the simulator for each action in those states to build an initial training dataset. Next, for the ``Na\"ive'' method, we keep on growing our dense dataset one state at a time while training and evaluating the decision tree every time, until the desired test accuracy is reached. For the ITRS method, we follow the same approach except we perform the rejection sampling. If ITRS rejects running the simulation for an action, it is not counted towards the dataset size. The results would be noisy if we stop when the desired test accuracy is reached the first time, as test accuracy is a random variable. Therefore, we wait until the desired accuracy is consistently achieved for at least 10 subsequent iterations, at which point, we record the dataset size. Recall that in our case, the simulator takes around 10 minutes for one run, therefore, the re-training, which takes around 1 minute, can easily be done while the simulator is being run for next data-point. \autoref{fig:performance} presents the dataset size required for various accuracy values. We observe that as the accuracy grows, the difference in the dataset size required by the two methods increase significantly in this case. For instance, in-order to reach an accuracy of $74\%$, the proposed method required only on average around 650 simulation runs, while the ``Naive'' method required on average around 2400 runs. Thus, we observed a reduction of around 68-74\% in training data, which amounts to the saving of around 265-300 hours of simulation time.
 
\subsection{Ablation study for the cost-sensitive classification}
We observe that for a certain box configuration, both sweep and pick actions might result in similar rewards, while the opposite is also true. Therefore, as discussed in \autoref{sec:training_with_sparse}, we used empirical rewards (1-cost) during the training of
the OSDT decision tree to penalize the mistakes made by the classifier based on the cost (cost-sensitive classifier). In this experiment we wanted to observe the effect of this choice. We train the decision trees using ITRS but once with a loss function that uses the empirical rewards and once with a loss function which uses misclassification as the loss function. Let us refer to the later as a ``baseline classifier''. For evaluation, we look at the percentage of average empirical rewards achieved by both the classifiers as compared to the maximum reward achievable by an oracle on the test set.  As expected, we observe that the cost-sensitive classifier resulted in a reward percentage of $83.78\%$ per action on average, while the baseline classifier resulted in a lower reward percentage of $81.74\%$ per action on average.

\subsection{In-depth analysis of ($\epsilon,\delta$)-\texttt{PAC} \texttt{NUSE}}
\begin{figure}[h!]
	\centering
	\includegraphics[width=\columnwidth]{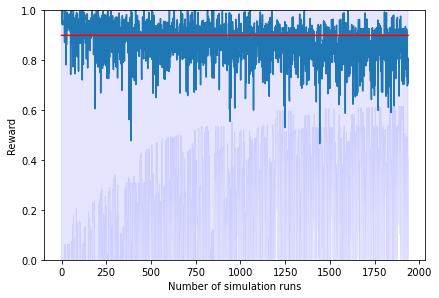}
	\caption{\small \textbf{Blue} curve shows what the algorithm believes the true reward for the best action is $\max_{j\in \chi} \hat{r}_{j,t}$, and the \textbf{light blue} region shows the confidence interval $[\max_{j\in \chi} \hat{r}_{j,t}-\max_i\epsilon_{i,t}, \max_{j\in \chi} \hat{r}_{j,t}+\max_i\epsilon_{i,t}]$. The \textbf{red} curve plots the true best reward.}
	\label{fig:nuse}
\end{figure}

In this experiment, we empirically evaluate the elimination criteria for ($\epsilon,\delta$)-\texttt{PAC} \texttt{NUSE}. We run the ITRS algorithm on a synthetic dataset containing 50 binary features, 3 actions, and with $\delta=0.05$, $\epsilon=0.35$. The synthetic dataset always had $0.9$ as the reward for the best action in each feature-space partition. \autoref{fig:nuse} shows that the true mean of $0.9$ is always captured by the empirical bounds.

\section{Conclusion and Future Work}
In this paper, we present Iterative Training with Rejection Sampling (ITRS) algorithm which iteratively trains an Optimal Decision Tree for Truck Unloading, a robotic Decision Making Problem while collecting additional data from simulation only when the new data would help train a better Decision Tree. We observe that with this method, we require around 68-74\% less simulator runs for the truck unloading problem, which amounts to saving about 265-300 hours of simulation time. The proposed method, provides better decision trees with lesser amount of data and hence increases the data efficiency which is crucial in cases like ours where it is very time consuming to obtain data.

Future work includes extending this algorithm to more general robotic decision making problems and reasoning about the sim-to-real gap when updating the decision tree, learned on simulation data, for a real-robot.

\section*{Appendix}
\label{sec:appendix}
\noindent \textbf{Proof:} \autoref{alg:nu-successive-elimination} is a ($0,\delta$)-\texttt{PAC} algorithm.\\
For an action $i$, let us define the event $\xi_{i,t} = \{ | \hat{r}_{i,t} - r_{i} | \leq \epsilon_t \}$. Using Hoeffding's inequality
\begin{align}
	\mathbb{P}(\xi_{i,t}^c) & \leq 2\text{exp}(-\tau_{i,t} \epsilon_{i,t}^2/2)
\end{align}
Taking $\epsilon_t =  \sqrt{ \frac{2}{\tau_t}\log(\frac{4t^2n}{\delta}) } $, we get
\begin{equation}
	\mathbb{P}(\xi_{i,t}^c) \leq \frac{\delta}{2t^2n}
\end{equation}
Thus, the new event $\xi$ where the event $\xi_{i,t}$ holds for all arms and at all times is $\xi = \bigcap\limits_{i=1}^{n} \bigcap\limits_{t=1}^{\infty} \xi_{i,t} $
\begin{align}
	\mathbb{P}(\xi^c) & \leq \sum\limits_{i=1}^{n} \sum\limits_{t=1}^{\infty} \frac{\delta}{2t^2n}\\
	& \leq \delta
\end{align}
Thus, $\mathbb{P}(\xi) \geq 1-\delta$. Now without loss of generality, let the actions be indexed in the decreasing order of their true expected rewards $r_1\geq r_2 \geq \ldots \geq r_n$. Thus, $a_1$ is the true best action. Let $\Delta_i = r_1 - r_{i}$ denote the gap, which is a non-negative quantity. Now, if event $\xi$ holds, the difference in the emperical mean rewards between any action $a_i \in \chi_t$ compared to the action $a_1$ is
\begin{align}
	\hat{r}_{i,t} - \hat{r}_{1,t} & = (\hat{r}_{i,t} - r_i) - (\hat{r}_{1,t} - r_1) - \Delta_i \nonumber\\
	& \leq \epsilon_{i,t} + \epsilon_{1,t} - \Delta_i\nonumber\\
	& \leq \epsilon_{i,t} + \epsilon_{1,t}\nonumber\\
	\hat{r}_{i,t} - \hat{r}_{1,t} & \leq \epsilon_{i,t} + \sideset{}{_j}{\max}\epsilon_{j,t}\label{eq:elim-crit}
\end{align}
In the last step we upper bound $\epsilon_{1,t}$ with $\sideset{}{_j}{\max}\epsilon_{j,t}$, as among all the actions, we don't know which one is the truly best. Further, the upper bound in \autoref{eq:elim-crit} holds true for all actions $a_i$ and for all time $t$ when $\xi$ holds. To be on the safest side and to not eliminate the best action (with high probability), we arrive at the following elimination rule:
\begin{equation}
	\sideset{}{_{j\in \chi}}{\max} \hat{r}_{j,t} - \hat{r}_{i,t} > 2\sideset{}{_j}{\max}\epsilon_{j,t}
\end{equation}
Thus, if we follow the above elimination rule, the best action will never (with high probability) be not included in $\chi_t$.

\bibliographystyle{IEEEtran}
\bibliography{references}

\end{document}